# JU_KS@SAIL_CodeMixed-2017: Sentiment Analysis for Indian Code Mixed Social Media Texts

**Kamal Sarkar**
Computer Science & Engineering Dept.,
Jadavpur University,
Kolkata- 700032, India
jukamal2001@yahoo.com

## Abstract

This paper reports about our work in the NLP Tool Contest @ICON-2017, shared task on Sentiment Analysis for Indian Languages (SAIL) (code mixed). To implement our system, we have used a machine learning algorithm called Multinomial Naïve Bayes trained using n-gram and SentiWordnet features. We have also used a small SentiWordnet for English and a small SentiWordnet for Bengali. But we have not used any SentiWordnet for Hindi language. We have tested our system on Hindi-English and Bengali-English code mixed social media data sets released for the contest. The performance of our system is very close to the best system participated in the contest. For both Bengali-English and Hindi-English runs, our system was ranked at the 3$^{rd}$ position out of all submitted runs and awarded the 3$^{rd}$ prize in the contest.

## 1 Introduction

Sentiment analysis deals with identification and classification of people's sentiments, attitudes and emotions from written texts. Recently, social media users ranging from celebrities to ordinary people, express their opinions, emotions or attitudes towards a broad range of topics on various platforms such as Twitter, Youtube, Wikipedia, Facebook etc. It is reported that there are more than 340 million tweets posted per day on Twitter. Users' postings on microblogging websites include real time messages about the real-life events, opinions on variety of topics and discussions of the current issues. The huge amount of dynamically changing various kinds of social media data coming from multiple sources transforms the social media data mining problem as big-data mining problem which is attracting the attention of different communities interested in analyzing its content. For example, some research works highlighted how social media content can be used to predict real-world outcomes, such as box-office revenues for movies (Asur and Huberman, 2010) or topic-based twitter sentiment analysis for stock prediction (Si et. al., ,2013) Among the applications, sentiment analysis (Agarwal et. al., 2011; Mohammad et. al, 2013) is one of the typical technique. Most of previous social media content analysis research deals with sentiment analysis of texts written in English. Indian social media texts written in Indian languages are gradually increasing on social media. Microblog can give a reflection of the reaction of the general public to social topics. The government organizations may be interested to collect and analyze feedback from the Indian people on policies to be framed by the Government of India. The various companies can also apply the same approach whenever they need to take decision regarding the products to be launched for the Indian market. The governments or the companies need to gather the public sentiment towards their policies or products and what are public opinion towards the various aspects of their policies or products. The police departments or homeland security departments may be interested to monitor social media for understanding public sentiment and moods, to detect in advance possible mass gatherings or protests. But, it is practically impossible to manually analyze the massive amount of social media data for real time decision making. So, there is a need for an automatic system which is capable of analyzing Indian social media texts for analyzing public sentiment in Indian social media texts.

This paper presents a Sentiment Analysis (SA) system for Indian social media texts which is developed for the SAIL 2017 contest@ICON2017, the goal of which is to perform sentiment polarity detection in code mixed Indian social media texts written in two different Indian languages, Hindi and Bengali mixed with English. We have participated for Hindi-English and Bengali-English language pairs

. The existing approaches to sentiment polarity detection problem can be classified as Lexicon-based approaches, Machine Learning (ML) based approaches and Hybrid approaches (Medhat et al,, 2014).

Lexicon-based approaches depend on a sentiment lexicon, a collection of known and precompiled sentiment terms. It is mainly divided into two approaches: Dictionary-based approaches and Corpus-based approaches. The work in (Minging and Bing, 2004; Kim and Hovy, 2004) represent the main strategy of Dictionary-based approaches. One of the Corpus-based approaches presented in (Hatzivassiloglou and McKeown, 1997) uses a list of seed opinion adjectives and imposed a set of linguistic constraints to specify additional adjective opinion words and their orientations.

Machine Learning (ML) based approaches are proven to be successful for SA tasks. ML based approach uses syntactic and/or linguistic features with the ML algorithms to solve the SA as a text classification problem (Medhat et al., 2014). Several ML algorithm have been used for solving sentiment classification problem such as Naive Bayes Classifier (NB) model (Hanhoon, Joon and Dongil, 2012), Support Vector Machines Classifiers (SVM) (Chien, and You-De, 2011) etc. SAIL 2015 contest conducted in conjunction with MIKE 2015 considered sentiment analysis of tweets in Indian languages. The systems that performed the best was based on machine learning algorithms (Sarkar 2018; Sarkar and Bhowmik, 2017; Patra et. al., 2015; Sarkar and Chakraborty, 2015).

This paper is organized as follows; a description of the training data was given in Section 2. The system description is given in section 3. Section 4 represents the experimental result and analysis of the result.

## 2 Training Data

The training data was provided by the organizer of SAIL 2017 in two different files – (1) one file containing Bengali-English code mixed social media texts which are basically tweets, (2) another file containing Hindi-English code mixed social media texts of similar kind. The training tweets were labeled with three labels- positive, negative and neutral. The details of training and test data are described later in this paper.

## 3 Methodology

We have viewed sentiment analysis as classification task and have used multinomial Naive Bayes for developing our proposed sentiment classification task. The description of our proposed system is given in the subsequent subsection.

### 3.1 Multinomial Naive Bayes Classifier

Naïve Bayes multinomial classifier (Kibriya,, Frank, Pfahringer and Holmes, 2005 ) computes class probabilities for a given text for purpose of classification. If $C$ is the set of classes and N is the size of a vocabulary, Naïve Bayes Multinomial classifier classifies a test tweet $t_i$ to the class which has the highest class membership probability $Pr(c|t_i)$ which can be computed as follows:

$$Pr(t_i|c) = \propto \prod_n Pr(w_n|c)^{f_{ni}},$$

Where: α is a constant.
$f_{ni}$ = the count of word $n$ in our test tweet $t_i$ and $Pr(w_n|c)$ = probability of word $n$ given class c, which is estimated from the training corpus of tweets as:

$$\widehat{Pr}(w_n|c) = \frac{1+Fr_{nc}}{N+\sum_{x=1}^{N}Fr_{xc}},$$

where :

$Fr_{xc}$ = count of word $x$ in all the training corpus of tweets in class c and N is the vocabulary size.

For implementing our sentiment analysis system, we have chosen the classifier "NaiveBayesMultinomialText", included in Weka. Weka is a machine learning toolkit (Hall et. al, 2009).

### 3.2 Feature Extraction

The feature extraction is a crucial component in any classification task. We have used n-gram features and SentiWordnet features. We have used a small SentiWordnet for English and a small SentiWordnet for Bengali. But we have not used any SentiWordnet for Hindi language. Our used Bengali SentiWordnet is a simple collection of 1700 positive words and 3750 negative words. It does not contain any neutral words. The English SentiWordnet consists of 2006 positive English words and 4783 negative English words. The English SentiWordNet does not contain any neutral words. Though the BengaliSentiWordnet is developed by us, the English SentiWordnet is created by collecting the positive and negative words used by Hu and Liu (2004) for mining and summarizing customer reviews. Bengali SentiWordnet was initially created in UTF 8 format using Bengali font. But this type of data is not suitable for sentiment analysis in code mixed tweet data supplied for SAIL 2017 contest because SAIL 2017 data set was released in Romanized format. So, we have transliterated all words in Bengali SentiwordNet into ITRANS Romanization format. The Bengali letter for which there was two possible transliterations in ITRANS format we have chosen the second one while transliterating all Bengali sentiment words. For example, the Bengali letter আ has two transliterations in ITRANS format: "A" and "aa", and we have chosen "aa" instead of "A" because we have observed that tweets contain "aa" more frequently in place of আ. Though some tweets also contain "a" in place of আ, we have not considered this issue in our current implementation.

Since tweets in the data sets released for the contest are language tagged (BN tag for Bengali word, HI tag for Hindi word and EN for English word), before feature extraction, we attach the language tag at the end of the corresponding word as <word_LanguageTag>. Before feature extraction, we also augment each token of the form <word_LanguageTag> with the sentiment tag for the corresponding word retrieved from the SentiWordNets. If the word is not found in SentiWordNet, then we use the tag "UNK". After this augmentation process, each word in a tweet takes a form: <<word_LanguageTag> senti-tag>. For example, the language tagged tweet "It's/EN a/EN darun/BN movie/EN" is processed as "It's_EN <UNK> a_EN <UNK> darun_BN <Positive> movie_EN <UNK>. Here <positive> is the sentiment tag taken from the Bengali SentiWordnet for the Bengali word "Darun" present in the input text.

Term frequency based bag-of-terms representation is used for creating vectors for the tweets. The size of the vector representing a tweet becomes *m* if the distinct numbers of *n*-grams (unigrams and bigrams in our case) in the training corpus is *m*. A tweet is represented as a vector of the form $<v_1, v_2, v_3, \ldots v_m>$, where the value of $v_i$ is the frequency of the i-th vocabulary word present in the tweet. The value of $v_i$ is set to 0, if the corresponding vocabulary word is not present in the tweet. Here vocabulary is created only from the training data. Since, before vector representation, the tweets are augmented with word level sentiment tagging using SentWordnet, word level sentiment tags are also considered as the tokens and added as the parts of unigram and bigram features used in our work. The vector obtained for each training tweet is labeled with the class of the corresponding tweet. The same procedure is also applied for vector representation of the test tweets.

### 3.3 System Architecture

We have developed our system by using the multinomial Naive Bayes classifier named "NaiveBayesMultinomialText" from WEKA. Weka is a machine learning toolkit consisting of a bunch of machine learning tools.

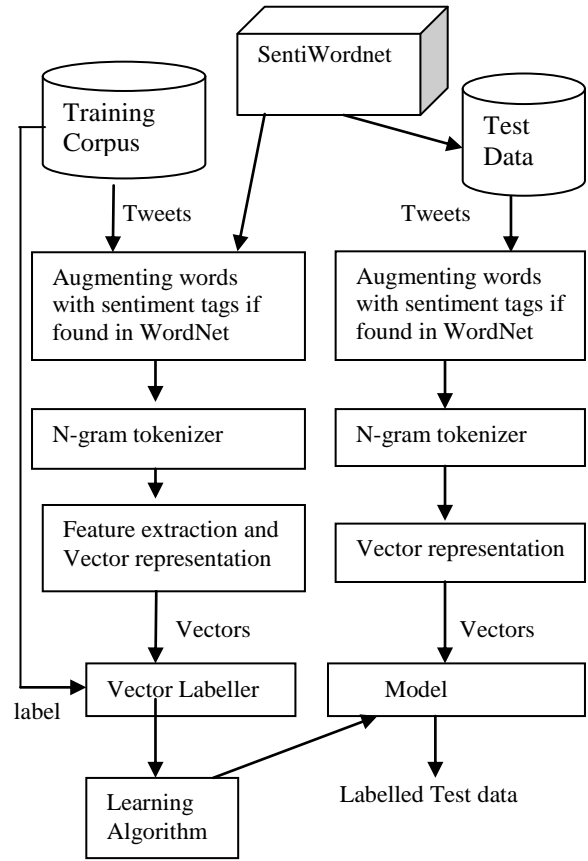

Figure 1. System Architecture for sentiment analysis of code mixed Indian social media texts

The training data and test data are submitted to the classifier. *N*-gram tokenizer present in WEKA has been used for tokenizing tweets. For obtaining the better results on BN_EN training data, we tune the parameters by performing 10-fold cross validation on the training data and we set the value of *n* to 2 (that is, unigram, bigram features) and frequency cut-off set to 2(this helps to remove the noisy features which does not occur in the corpus at least twice). For HN-EN data, we have set the value of *n* to 2 (that is, unigram, bigram features) and frequency cut-off set to 1 for the best results on the training data. As discussed earlier in this paper, term frequency based bag-of-terms representation is used for creating vectors for the tweets. The system architectures that we used for our sentiment analysis task is shown in figure 1.

## 4 Evaluation and Results

We submitted one run for each language pair- (1) one run for Bengali-English pair and (2) one run for Hindi-English pair.

| Language | Training Data (number of tweets) | | | Test Data (number of tweets) |
|---|---|---|---|---|
| | Positive | Negative | Neutral | |
| Bengali-English pair | 1000 | 1000 | 500 | 3038 |
| Hindi-English pair | 4064 | 2972 | 5900 | 5525 |

Table 1. Description of the data sets

The classification outputs produced on the test data by our developed system was sent to the organizers for evaluation. The results of our submitted runs were announced and reported along with the results of the runs submitted by other participating teams.

The outputs of the participating systems were evaluated using precision, Recall, f-score. The organizers have calculated F-score using the formula: f-score = (fpos +fneg+fneu)/3, where fpos, fneg and fneu are f-score for positive, negative and neutral class respectively.

The description of the training and test data is shown in the Table 1. Table 2 shows the performances of our developed sentiment analysis system on the test data for the two language pairs namely Bengali-English (BN-EN) and Hindi-English (HN-EN). From table 2 it can be seen that the overall f-score of our system for Bengali-English pair and Hindi-English pair 0.504 and 0.562 respectively.

| Overall Performance | | | |
|---|---|---|---|
| Lang. pair | Precision (P) | Recall (R) | F-score |
| BN-EN | 0.606 | 0.524 | 0.504 |
| HI-EN | 0.579 | 0.556 | 0.562 |

Table 2. Performance of our sentiment analysis system on the test data for Bengali-English(BN-EN) and Hindi-English (HI-EN) language pair

### 4.1 Performance Comparison

In this section, we describe the results for the SAIL contest 2017@ICON 2017 in which a total of 9 systems participated for BN-EN language pair and 14 systems participated for HN-EN language pair. The official results announced by the organizers for BN-EN and HN-EN language pairs are shown in Table 3 and Table 4 respectively. We have shown in the tables the overall performances of the systems participated in the contest.

| Overall performance on BN-EN test data | | | |
|---|---|---|---|
| SystemID | Run ID | P | R | F-Score |
| IIIT-NBP | 1 | 0.552 | 0.531 | 0.524 |
| IIIT-NBP | 2 | 0.551 | 0.534 | 0.526 |
| NLP-CEN-AMRITA-RBG | 1 | 0.517 | 0.516 | 0.513 |
| **JU_KS** | **1** | **0.606** | **0.524** | **0.504** |
| CFILT | 1 | 0.528 | 0.476 | 0.455 |
| CFILT | 2 | 0.538 | 0.478 | 0.447 |
| Random Baseline | | 0.342 | 0.343 | 0.339 |
| AMRITA_CEN | 1 | 0.322 | 0.34 | 0.318 |
| CEN@ Amrita | 1 | 0.23 | 0.309 | 0.258 |
| SVNIT | 1 | 0.136 | 0.333 | 0.193 |

Table 3. The official results announced for sentiment analysis of Bengali-English code mixed Indian social media texts

We can see from Table 3, performance of our system JU_KS is comparable to the best performing systems contested for sentiment analysis task for BN-EN data. We can also observe from the table that overall precision of our system JU_KS is better than the best system IIIT-NBP and the overall recall of our system is also very close to the best system though the overall f-measure of our system is lower than the best system. This is due to how the organizers have calculated f-score of the overall system.

| Overall performance on HI-EN data | | | | |
|---|---|---|---|---|
| SystemID | Run ID | P | R | F-Score |
| IIIT-NBP | 2 | 0.597 | 0.56 | 0.569 |
| BIT Mesra Ambuj | 1 | 0.573 | 0.558 | 0.564 |
| BIT Mesra mbuj | 2 | 0.573 | 0.559 | 0.564 |
| **JU-KS** | **1** | **0.579** | **0.556** | **0.562** |
| IIIT-NBP | 1 | 0.607 | 0.547 | 0.557 |
| NLP-CEN-AMRITA-RBG | 1 | 0.589 | 0.54 | 0.55 |
| CFILT | 2 | 0.591 | 0.516 | 0.524 |
| CFILT | 1 | 0.575 | 0.508 | 0.514 |
| Harsh | 2 | 0.464 | 0.462 | 0.461 |
| Harsh | 1 | 0.46 | 0.46 | 0.459 |
| Random Baseline | | 0.337 | 0.338 | 0.331 |
| CEN@Amrita | 2 | 0.275 | 0.354 | 0.309 |
| CEN@Amrita | 1 | 0.43 | 0.339 | 0.3 |
| SVNIT | 1 | 0.105 | 0.333 | 0.16 |
| SVNIT | 2 | 0.105 | 0.333 | 0.16 |

Table 4. The official results announced for sentiment analysis of Hindi-English code mixed Indian social media texts

As we can see from the table 4, performance of our system JU_KS is comparable to the best performing systems contested for sentiment analysis task for HN-EN data. The overall performance of our proposed system on HN-EN test data is close (it only differs in the last digit) to the best system IIIT-NBP. Table 4 also shows that the overall precision of our system is better than the second best system, *BIT Mesra Ambuj* and the overall recall is very close to the second best system though overall f-score is lower than that obtained by the system *BIT Mesra Ambuj*.

## 5 Conclusion

This paper describes a sentiment analysis system for sentiment analysis of code mixed social media texts in Indian languages (Hindi-English and Bengali-English pairs). Two different runs have been performed: one run for Bengali-English language pair and another run for Hindi-English language pair. The official results published by the organizers show that performance of our system is comparable to other systems that perform the best in the contest. The further improvement of our system is possible by adding more words to the SentiWordnet used for implementing our system. Other machine learning algorithms like SVM and deep learning can also be applied for improving system performance.